\documentclass[conference]{IEEEtran}
\IEEEoverridecommandlockouts
\usepackage{cite}
\usepackage{amsmath,amssymb,amsfonts}
\usepackage{algorithmic}
\usepackage{graphicx}
\usepackage{textcomp}
\usepackage{xcolor}
\usepackage{graphicx,subfigure,amsmath,amsfonts,amssymb,algorithm,algorithmic,cite,multirow,rotating,multicol,booktabs,threeparttable,bm}
\def\BibTeX{{\rm B\kern-.05em{\sc i\kern-.025em b}\kern-.08em
    T\kern-.1667em\lower.7ex\hbox{E}\kern-.125emX}}
\begin{document}

\title{Curse of Dimensionality for TSK Fuzzy Neural Networks: Explanation and Solutions}

\author{Yuqi Cui, Dongrui Wu and Yifan Xu \\
\IEEEauthorblockA{\textit{School of Artificial Intelligence and Automation} \\
\textit{Huazhong University of Science and Technology}\\
Wuhan, China \\
\{yqcui, drwu, yfxu\}@hust.edu.cn}
}

\maketitle

\begin{abstract}
  Takagi-Sugeno-Kang (TSK) fuzzy system with Gaussian membership functions (MFs) is one of the most widely used fuzzy systems in machine learning. However, it usually has difficulty handling high-dimensional datasets. This paper explores why TSK fuzzy systems with Gaussian MFs may fail on high-dimensional inputs. After transforming defuzzification to an equivalent form of \emph{softmax} function, we find that the poor performance is due to the saturation of \emph{softmax}. We show that two defuzzification operations, LogTSK and HTSK, the latter of which is first proposed in this paper, can avoid the saturation. Experimental results on datasets with various dimensionalities validated our analysis and demonstrated the effectiveness of LogTSK and HTSK.
\end{abstract}

\begin{IEEEkeywords}
  Mini-batch gradient descent, fuzzy neural network, high-dimensional TSK fuzzy system, HTSK, LogTSK
\end{IEEEkeywords}

\section{Introduction}

Takagi-Sugeno-Kang (TSK) fuzzy systems~\cite{takagi1985fuzzy} have achieved great success in numerous machine learning applications, including both classification and regression. Since a TSK fuzzy system is equivalent to a five layer neural network~\cite{jang1993anfis,drwuTSK2020}, it is also known as TSK fuzzy neural network. 

Fuzzy clustering~\cite{xu2019concise,deng2010scalable,cui2020supervised} and evolutionary algorithms~\cite{shi1999implementation,wu2006genetic} have been used to determine the parameters of TSK fuzzy systems on small datasets. However, their computational cost may be too high for big data. Inspired by its great success in deep learning~\cite{sutskever2013importance,kingma2014adam,luo2019adaptive}, mini-batch gradient descent (MBGD) based optimization was recently proposed for training TSK fuzzy systems~\cite{wu2019optimize,cui2020optimize}.

Traditional optimization algorithms for TSK fuzzy systems use grid partition to partition the input space into different fuzzy regions, whose number grows exponentially with the input dimensionality. A more popular and flexible way is clustering-based partition, e.g., fuzzy c-means (FCM)~\cite{bezdek1984fcm}, EWFCM~\cite{zhou2016fuzzy}, ESSC~\cite{deng2010enhanced,xu2019concise} and SESSC~\cite{cui2020supervised}, in which the fuzzy sets in different rules are independent and optimized separately.

Although the combination of MBGD-based optimization and clustering-based rule partition can handle the problem of optimizing antecedent parameters on high-dimensional datasets, TSK fuzzy systems still have difficulty achieving acceptable performance. The main reason is the curse of dimensionality, which affects all machine learning models. When the input dimensionality is high, the distances between data points become very similar~\cite{houle2010can}. TSK fuzzy systems usually use distance based approaches to compute membership grades, so when it comes to high-dimensional datasets, the fuzzy partitions may collapse. For instance, FCM is known to have trouble handling high-dimensional datasets, because the membership grades of different clusters become similar, leading all centers to move to the center of the gravity~\cite{winkler2011fuzzy}.

Most previous works used feature selection or dimensionality reduction to cope with high-dimensionality. Model-agnostic feature selection or dimensionality reduction algorithms, such as Relief~\cite{urbanowicz2018relief} and principal component analysis (PCA)~\cite{wold1987principal,gu2020distilling}, can filter the features before feeding them into TSK models. Neural networks pre-trained on large datasets can also be used as feature extractor to generate high-level features with low dimensionality~\cite{deng2016hierarchical,du2020tsk}.

There are also approaches to select the fuzzy sets in each rule so that rules may have different numbers of antecedents. For instance, Alcala-Fdez \emph{et al.} proposed an association analysis based algorithm to select the most representative patterns as rules~\cite{alcala2011fuzzy}. Cózar \emph{et al.} further improved it by proposing a local search algorithm to select the optimal fuzzy regions~\cite{cozar2018learning}. Xu \emph{et al.} proposed to use the attribute weights learned by soft subspace fuzzy clustering to remove fuzzy sets with low weights to build a concise TSK fuzzy system~\cite{xu2019concise}. However, there were few approaches that directly train TSK models on high-dimensional datasets.

Our previous experiments found that when using MBGD-based optimization, the initialization of the standard deviation of Gaussian membership functions (MFs), $\sigma$, is very important for high-dimensional datasets, and larger $\sigma$ may lead to better performance. In this paper, we demonstrate that this improvement is due to the reduction of saturation caused by the increase of dimensionality. Furthermore, we validate two convenient approaches to accommodate saturation.

Our main contributions are:
  \begin{itemize}
    \item To the best of our knowledge, we are the first to discover that the curse of dimensionality in TSK modeling is due to the saturation of the \emph{softmax} function. As a result, there exists an upper bound on the number of rules that each input can fire. Furthermore, the loss landscape of a saturated TSK system is more rugged, leading to worse generalization.
    \item We demonstrate that the initialization of $\delta$ should be correlated with the input dimensionality to avoid saturation. Based on this, we propose a high-dimensional TSK (HTSK) algorithm, which can be viewed as a new defuzzification operation or initialization strategy.
    \item We validate LogTSK \cite{du2020tsk} and our proposed HTSK on datasets with a large range of dimensionality. The results indicate that HTSK and LogTSK can not only avoid saturation, but also are more accurate and more robust than the vanilla TSK algorithm with simple initialization.
  \end{itemize}

The remainder of this paper is organized as follows: Section~\ref{sec:tsk} introduces TSK fuzzy systems and the saturation phenomenon on high-dimensional datasets. Section~\ref{sec:high_dim_tsk} introduces the details of LogTSK and our proposed HTSK. Section~\ref{sec:result} validates the performances of LogTSK and HTSK on datasets with various dimensionality. Section~\ref{sec:conclusion} draws conclusions.

\section{Traditional TSK Fuzzy Systems on High-dimensional Datasets}\label{sec:tsk}

This section introduces the details of TSK fuzzy system with Gaussian MF \cite{drwuEAAI2019}, the equivalence between defuzzification and \emph{softmax} function, and the saturation phenomenon of \emph{softmax} on high-dimensional datasets.

\subsection{TSK Fuzzy Systems}\label{ssec:tsk}

Let the training dataset be $\mathcal{D}=\{\bm{x}_n, y_n\}_{n=1}^N$, in which $\bm{x}_n=[x_{n,1},...,x_{n,D}]^T\in \mathbb{R}^{D\times 1}$ is a $D$-dimensional feature vector, and $y_n\in\{1,2,...,C\}$ can be the corresponding class label for a $C$-class classification problem, or $y_n\in \mathbb{R}$ for a regression problem.

Suppose a $D$-input single-output TSK fuzzy system has $R$ rules. The $r$-th rule can be represented as:
    \begin{align}\label{eq:rule_r}\begin{aligned}
      \text{Rule}_r: &\text{IF}~x_1~\text{is}~X_{r,1}~\text{and}~\cdots~\text{and}~x_D~\text{is}~X_{r,D},\\
      &\text{Then}~y_r(x)=b_{r,0}+\sum_{d=1}^Db_{r,d}x_d,
    \end{aligned}
    \end{align}
where $X_{r,d}$ ($r=1,...,R$; $d=1,...,D$) is the MF for the $d$-th attribute in the $r$-th rule, and $b_{r,d}$, $d=0,...,D$, are the consequent parameters. Note that here we only take single-output TSK fuzzy systems as an example, but the phenomenon and conclusion can also be extended to multi-output TSK systems.

Consider Gaussian MFs. The membership degree $\mu$ of $x_d$ on $X_{r,d}$ is:
    \begin{align}\label{eq:mu_gmf}
      \mu_{X_{r,d}}(x_d)=\exp\left(-\frac{(x_d-m_{r,d})^2}{2\sigma_{r,d}^2}\right),
    \end{align}
where $m_{r,d}$ and $\sigma_{r,d}$ are the center and the standard deviation of the Gaussian MF $X_{r,d}$, respectively.

The final output of the TSK fuzzy system is:
    \begin{align}\label{eq:tsk_output}
      y(\bm{x})=\frac{\sum_{r=1}^Rf_r(\bm{x})y_r(\bm{x})}{\sum_{i=1}^Rf_{i}(\bm{x})},
    \end{align}
where
    \begin{align}\label{eq:tsk_fr}
      f_r(\bm{x}) = \prod_{d=1}^{D}\mu_{X_{r,d}}(x_d)=\exp\left(-\sum_{d=1}^{D}\frac{(x_d-m_{r,d})^2}{2\sigma_{r,d}^2}\right)
    \end{align}
is the firing level of Rule $r$. We can also re-write (\ref{eq:tsk_output}) as:
    \begin{align}\label{eq:tsk_output2}
      y(\bm{x})=\sum_{r=1}^{R}\overline{f}_r(\bm{x})y_r(\bm{x}),
    \end{align}
where
    \begin{align}\label{eq:tsk_norm_frs}
      \overline{f}_r(\bm{x}) = \frac{f_r(\bm{x})}{\sum_{i=1}^Rf_i(\bm{x})}
    \end{align}
is the normalized firing level of Rule $r$. (\ref{eq:tsk_output2}) is the defuzzification operation of TSK fuzzy systems.


In this paper, we use $k$-means clustering to initialize the antecedent parameters $m_{r,d}$, and MBGD to optimize the parameters $b_{r,d}$, $m_{r,d}$ and $\sigma_{r,d}$. More specifically, we run $k$-means clustering and assign the $R$ cluster centers to $m_{r,d}$ as the centers of the rules. We use different initializations of $\sigma_{r,d}$ to validate their influence on the performance of TSK models on high-dimensional datasets. He initialization \cite{he2015delving} is used for the consequent parameters.

\subsection{TSK Fuzzy Systems on High-Dimensional Datasets}\label{ssec:tsk_high_dim}

When using Gaussian MFs and the product $t$-norm, we can re-write (\ref{eq:tsk_norm_frs}) as:
    \begin{align}\label{eq:tsk_softmax_pre}\begin{aligned}
      \overline{f}_r(\bm{x}) &= \frac{f_r(\bm{x})}{\sum_{i=1}^Rf_i(\bm{x})}\\ &=\frac{\exp\left(-\sum_{d=1}^{D}\frac{(x_d-m_{r,d})^2}{2\sigma_{r,d}^2}\right)}
      {\sum_{i=1}^{R}\exp\left(-\sum_{d=1}^{D}\frac{(x_d-m_{i,d})^2}{2\sigma_{i,d}^2}\right)}.
    \end{aligned}
    \end{align}
Replacing $-\sum_{d=1}^{D}\frac{(x_d-m_{r,d})^2}{2\sigma_{r,d}^2}$ with $Z_{r}$, we can observe that $\overline{f}_r$ is a typical \emph{softmax} function:
    \begin{align}\label{eq:tsk_softmax}
      \overline{f}_r(\bm{x})=\frac{\exp(Z_r)}{\sum_{i=1}^R\exp(Z_i)},
    \end{align}
where $Z_r < 0$ $\forall \bm{x}$. We can also show that, as the dimensionality increases, $Z_r$ decreases, which causes the saturation of \emph{softmax}~\cite{chen2017noisy}.

Let $\bm{Z}=[Z_1, ..., Z_R]$ and $\bm{\overline{f}}=[\overline{f}_1, ..., \overline{f}_R]$. In a three-rule TSK fuzzy system for low-dimensional task, if $\bm{Z}=[-0.1, -0.5, -0.3]$, then $\overline{\bm{f}} = [0.4018, 0.2693, 0.3289]$. As the dimensionality increases, $\bm{Z}$ may increase to, for example, $[-10, -50, -30]$, and then $\overline{\bm{f}}=[1,4\times10^{-18}, 2\times10^{-9}]$, which means the final prediction is dominated by one rule. In other words, $\bm{\overline{f}}$ in (\ref{eq:tsk_softmax}) with high-dimensional inputs tends to only give non-zero firing level to the rule with the maximum $Z_r$.

In order to avoid numeric underflow, we compute the normalized firing level by a common trick:
    \begin{align}
      \overline{f}_r(\bm{x}) = \frac{\exp(Z_r-Z_{\max})}{\sum_{i=1}^R\exp(Z_i - Z_{\max})},
    \end{align}
where $Z_{\max}=\max(Z_1,...,Z_R)$. In this paper, we consider that a rule is fired by $\bm{x}$ when the corresponding normalized firing level $\overline{f}_r(\bm{x}) > 10^{-4}$.

\begin{figure}[htpb]
      \centering
       \subfigure[]{\includegraphics[clip,width=0.48\textwidth]{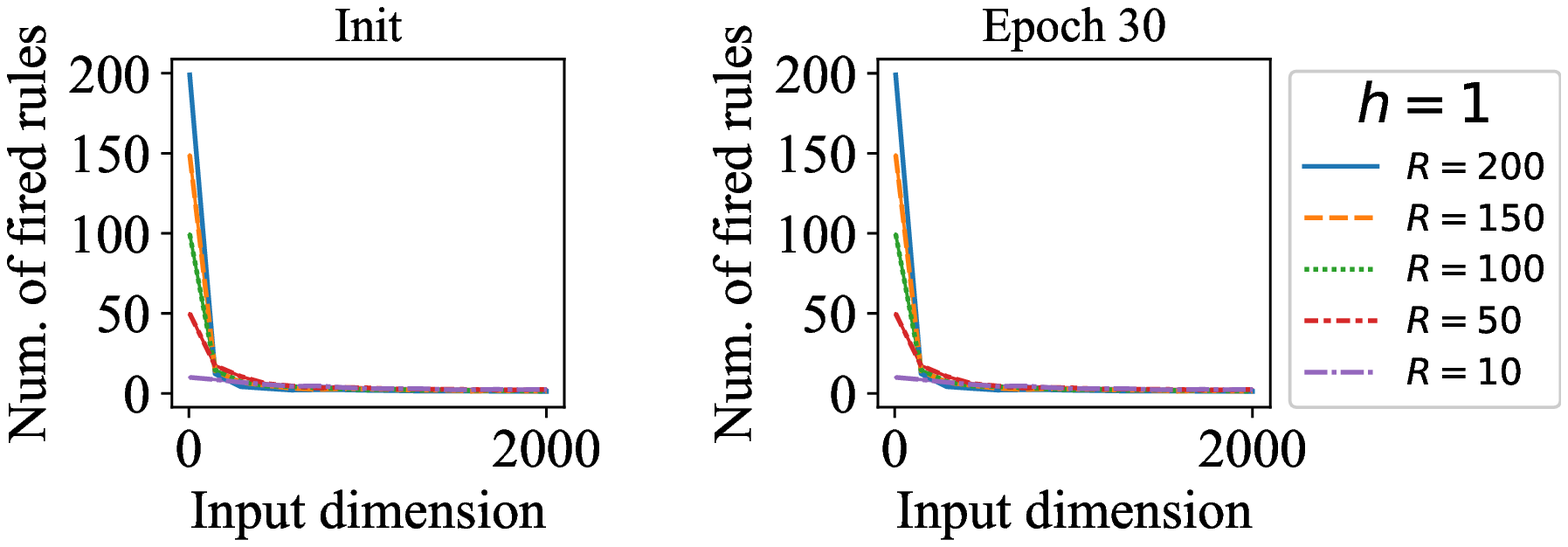}}
       \subfigure[]{\includegraphics[clip,width=0.48\textwidth]{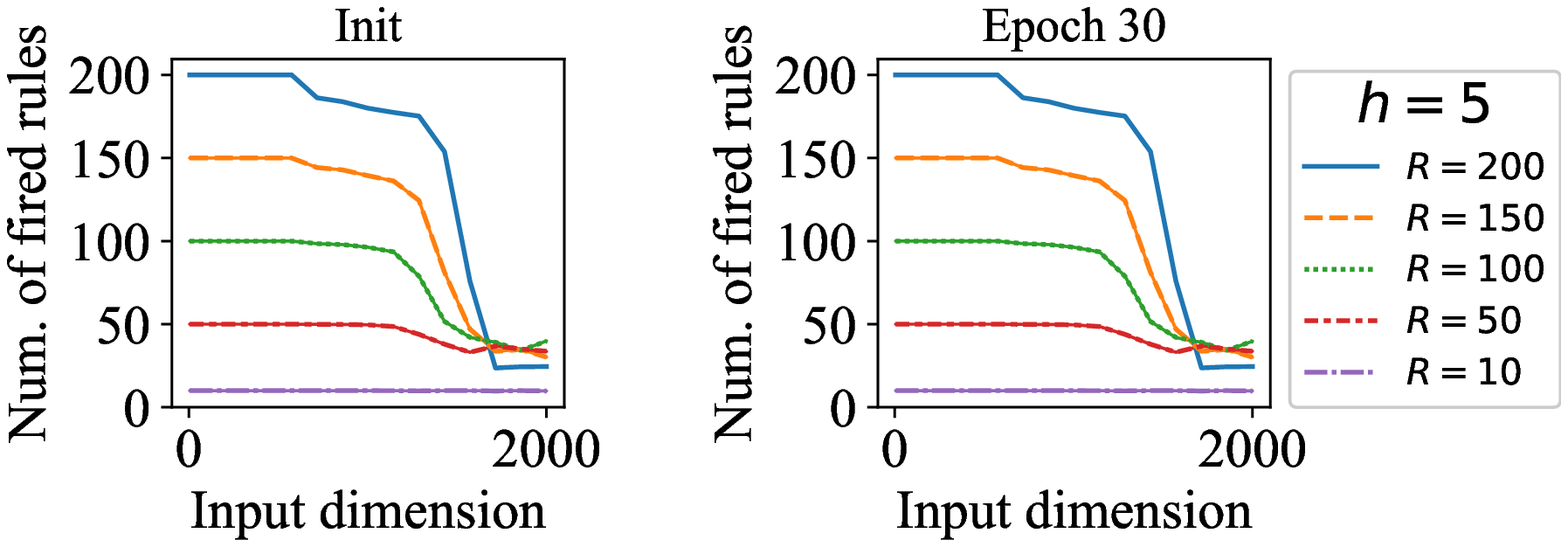}}
      \subfigure[]{\includegraphics[clip,width=0.48\textwidth]{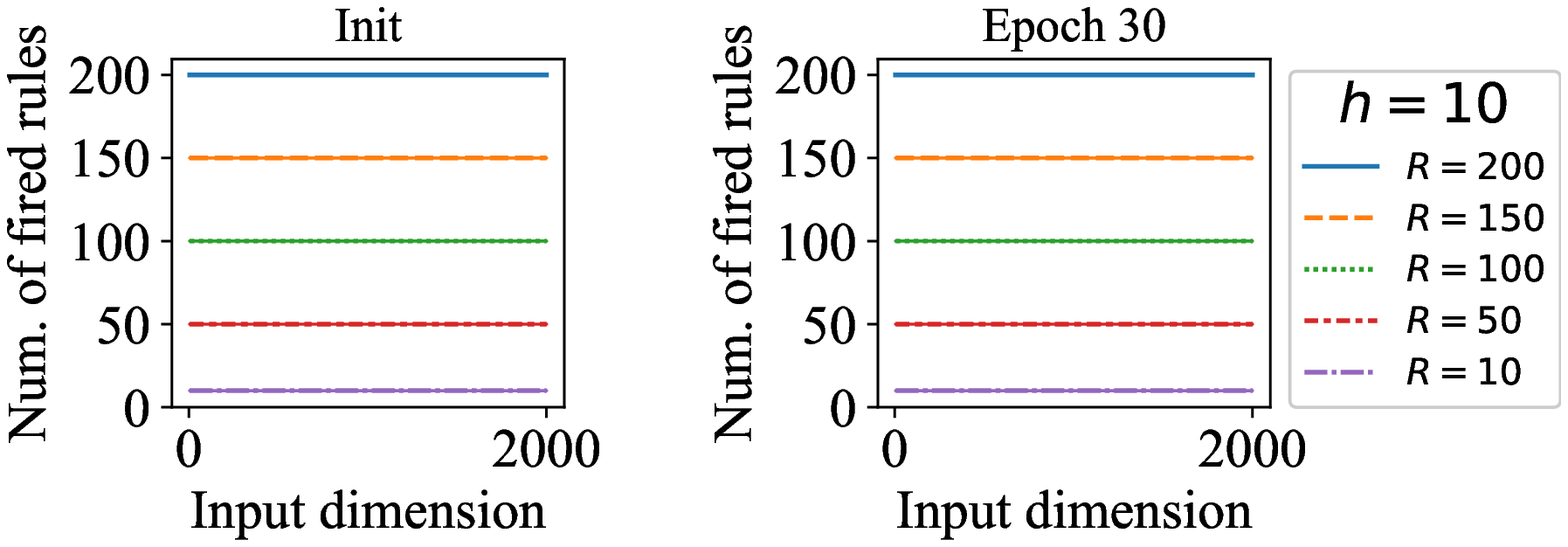}}
      \subfigure[]{\includegraphics[clip,width=0.48\textwidth]{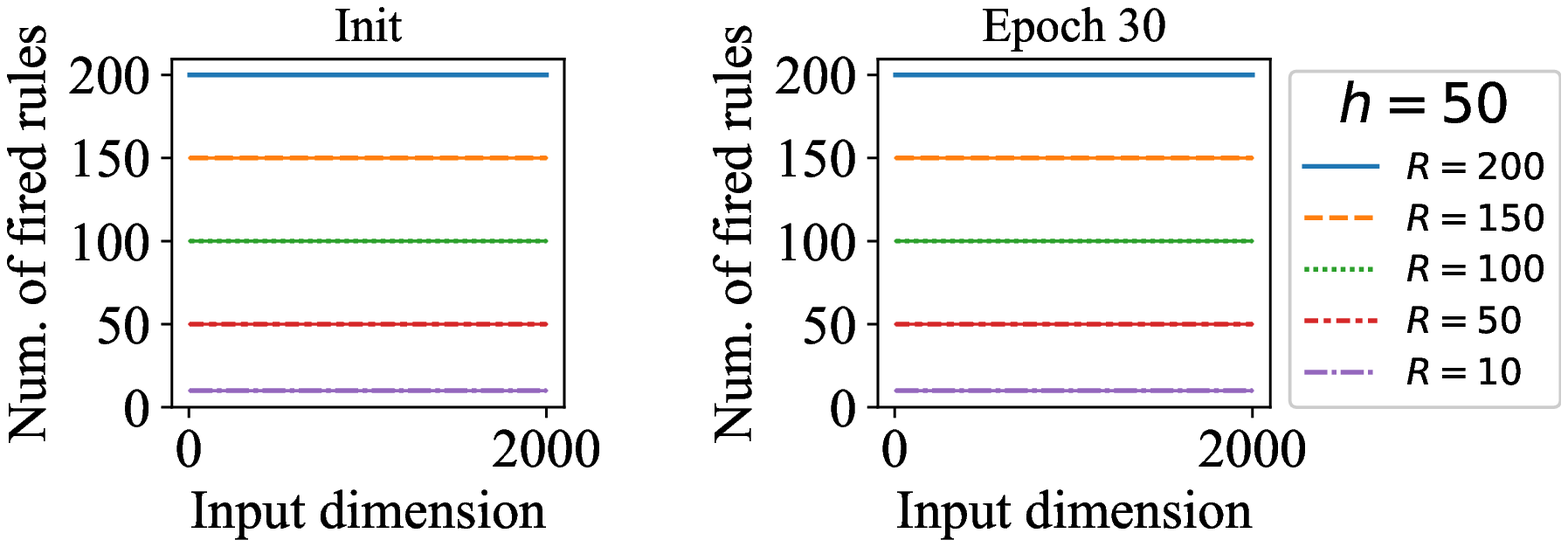}}
      \caption{The average number of fired rules versus the input dimensionality on randomly generated datasets. $\sigma$ of TSK models is initialized by Gaussian distribution $\sigma\sim\mathcal{N}(h,0.2), h=1,5,10,50$. The first and second columns represent the model before training and after 30 epochs of training, respectively.}\label{fig:dim_frs}
    \end{figure}

We generate a two-class toy dataset following Gaussian distribution $x_i\sim\mathcal{N}(0,1)$ with the dimensionality varying from 5 to 2,000 for pilot experiments. The labels are also generated randomly. We initialize $\sigma$ following Gaussian distribution $\sigma\sim\mathcal{N}(h,0.2)$, $h=1,5,10,50$, and train TSK models with different $R$ for 30 epochs. The number of rules fired by each input with different dimensionality at different training epochs is shown in Fig.~\ref{fig:dim_frs}. The number of fired rules decreases rapidly as the dimensionality increases when $h=1$. For a particular input dimensionality $D$, there exists an upper bound of the number of fired rules, i.e., larger $R$ would not always increase the number of fired rules. Increasing $h$ can mitigate the saturation phenomenon to a certain extent and increase the upper bound of the number of fired rules.

\begin{figure}[htpb]
      \centering
      \includegraphics[width=0.9\columnwidth,clip]{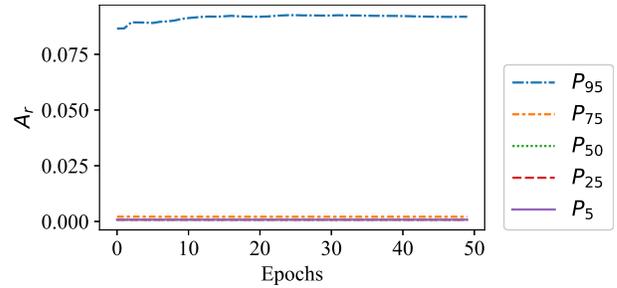}
      \caption{Different percentiles of $A_r$, $r=1,...,R$ versus the training epochs.} \label{fig:tsk_fired_by_dataset}
    \end{figure}

Although each high-dimensional input feature vector can only fire a limited number of rules due to saturation, different inputs may fire different subsets of rules, which means evert rule is useful to the final predictions. We compute the average normalized firing level of the $r$-th rule during training  by:
    \begin{align}\label{eq:ave_fr}
      A_r=\frac{1}{N}\sum_{n=1}^N\overline{f}_r(\bm{x}_n).
    \end{align}
We train TSK models with 60 rules and computed the 5\%-95\% percentiles of $A_r$, $r=1,...,R$, during training on dataset Books from Amazon product review datasets. The details of this dataset will be introduced in Section~\ref{sec:dataset}. We repeat the experiments ten times and show the average results in Fig.~\ref{fig:tsk_fired_by_dataset}. Except a small number of rules with high $A_r$, most rules barely contribute to the prediction. This phenomenon doesn't change as the training goes on.

\subsection{Enhance the Performance of TSK Fuzzy Systems on High-Dimensional Datasets}

The easiest way to mitigate saturation is to increase the scale of $\sigma$. As indicated by (\ref{eq:tsk_softmax_pre}) and (\ref{eq:tsk_softmax}), increasing the scale of $\sigma$ also increases the value of $Z_r$ to avoid saturation. Similar tricks have already been used for training TSK models with fuzzy clustering algorithms, such as FCM~\cite{bezdek1984fcm}, ESSC~\cite{deng2010enhanced} and SESSC~\cite{cui2020supervised}. The parameter $\sigma$ is computed by:
    \begin{align}\label{eq:fcm_sigma}
      \sigma_{r,d} = h\left[\sum_{n=1}^{N}U_{n,r}(x_{n,d}-V_{r,d})\left/\sum_{i=1}^{N}U_{i,r}\right.\right]^{1/2},
    \end{align}
where $U_{n,r}$ is the membership grade of $\bm{x}_n$ in the $r$-th cluster, $V_{r,\cdot}$ is the center of the $r$-th cluster, and $h$ is used to adjust the scale of $\sigma_{r,d}$. The larger $h$ is, the smaller $|Z_r|$ is. For MBGD-based optimization, we can directly initialize $\sigma$ with a proper value to avoid saturation in training. However, the proper scale parameter $h$ for $\sigma$ usually depends on the characteristics of the task, which requires trial-and-error, or time-consuming cross-validation.

%

A better way is to use a scaling factor depending on the dimensionality $D$ to constrain the range of $|Z_r|$. A similar approach is used in the Transformer~\cite{vaswani2017attention}, in which a scaling factor $1/\sqrt{d_k}$ is used to constrain the value of $QK^T$. When the distribution of the constrained $Z_r$ is no longer depending on the dimensionality $D$, all we have to do is to choose one proper initialization range of $\sigma$ suitable for most datasets. Alternatively, we can use other normalization approaches which are insensitive to the scale of $Z_r$. For instance, we can replace the defuzzification by
\begin{align}
\overline{f}_r(Z_r)=\frac{Z_r}{\|[Z_1, ..., Z_R]\|_1}
\end{align}
or
\begin{align}
\overline{f}_r(Z_r)=\frac{Z_r}{\|[Z_1, ..., Z_R]\|_2},
\end{align}
so that $\overline{f}_r(Z_r) = \overline{f}_r(hZ_r)$ $\forall h > 0$.

\section{Defuzzification for High-Dimensional Problems}\label{sec:high_dim_tsk}

This section introduces LogTSK proposed by Du \emph{et al.}~\cite{du2020tsk} and our proposed HTSK. Both are suitable for high-dimensional problems.

\subsection{LogTSK}\label{ssec:logtsk}

Recently, an algorithm called TCRFN was proposed for predicting driver's fatigue using the combination of convolutional neural network (CNN) and recurrent TSK fuzzy system~\cite{du2020tsk}. Within TCRFN, a logarithm transformation of $f_r$ was proposed to \emph{``amplify the small differences on firing levels''}. The firing level and normalized firing level of the $r$-th rule in TCRFN are:
    \begin{align}\label{eq:logtsk}
    \begin{aligned}
        &f_r^{log} = -\frac{1}{\log f_r}=\frac{1}{\sum_{d=1}^{D}\frac{(x_d-m_{r,d})^2}{2\sigma_{r,d}^2}}\\
        &\overline{f}_r^{log} = \frac{f_r^{log}}{\sum_{i=1}^{R}f_r^{log}}.
    \end{aligned}
    \end{align}
The final output is:
    \begin{align}\label{eq:logtsk_output}
      y(\bm{x}) = \sum_{r=1}^{R}\overline{f}_r^{log}(\bm{x})y_r(\bm{x}).
    \end{align}
We denote the TSK fuzzy system with this log-transformed defuzzification LogTSK in this paper.
Substituting $Z_r$ into (\ref{eq:logtsk}) gives
    \begin{align}\label{eq:logtsk_defuzz}
      \overline{f}_r^{log}=\frac{-1/Z_r}{-\sum_{i=1}^{R}1/Z_i}=\frac{-1/Z_r}{\|[1/Z_1,...,1/Z_R]\|_1},
    \end{align}
i.e., LogTSK avoids the saturation by changing the normalization from \emph{softmax} to $L_1$ normalization. Since $L_1$ normalization is not affected by the scale of $Z_r$, LogTSK can make TSK fuzzy systems trainable on high-dimensional datasets.

\subsection{Our Proposed HTSK}\label{ssec:htsk}

We propose a simple but very effective approach, HTSK (high-dimensional TSK), to enable TSK fuzzy systems to deal with datasets with any dimensionality, by avoiding the saturation in (\ref{eq:tsk_softmax}). HTSK  constrains the scale of $|Z_r|$ by simply changing the sum operator in $Z_r$ to average:
    \begin{align}\label{eq:zr1}
      Z_r'=-\frac{1}{D}\sum_{d=1}^{D}\frac{(x_d-m_{r,d})^2}{2\sigma_{r,d}^2}.
    \end{align}
We can understand this transformation from the perspective of defuzzification. (\ref{eq:tsk_output2}) can be rewritten as:
    \begin{align}\label{eq:htsk_output2}
      y(\bm{x})=\sum_{r=1}^{R}\overline{f}_r'(\bm{x})y_r(\bm{x}),
    \end{align}
    where
    \begin{align}\label{eq:htsk_norm_frs}
      \overline{f}_r'(\bm{x}) = \frac{f_r(\bm{x})^{1/D}}{\sum_{i=1}^{R}f_i(\bm{x})^{1/D}}=\frac{\exp(Z_r')}{\sum_{i=1}^R\exp(Z_i')}.
    \end{align}
In this way, the scale of $|Z_r'|$ no longer depends on the dimensionality $D$. Even in a very high dimensional space, if the input feature vectors are properly pre-processed (z-score or zero-one normalization, etc.), we can still guarantee the stability of HTSK.

HTSK is equivalent to adaptively increasing $\sigma$ $\sqrt{D}$ times in the vanilla TSK, i.e., the initialization of $\sigma$ should be correlated with the input dimensionality. The vanilla TSK fuzzy system is a special case of HTSK when setting $D=1$.

\section{Results}\label{sec:result}

In this section, we validate the performances of LogTSK and our proposed HTSK on multiple datasets with varying size and input dimensionality.

\subsection{Datasets}\label{sec:dataset}

Fourteen datasets with the dimensionality $D$ varying from 10 to 4,955 were used. Their details are summarized in Table~\ref{tab:dataset}. For FashionMNIST and MNIST, we used the official training-test partition. For other datasets, we randomly selected 70\% samples for training and the remaining 30\% for test.

    \begin{table}[htpb]\centering    \setlength{\tabcolsep}{0.3mm}
    \caption{Summary of the fourteen datasets.}\label{tab:dataset}
    \begin{threeparttable}
        \begin{tabular}{cccc}\hline
        Dataset                                   & Num. of   features & Num. of samples & Num. of classes \\\hline
        Vowel\tnote{1}           & 10                 & 990             & 11              \\
        Vehicle\tnote{1}         & 18                 & 596             & 4               \\
        Biodeg\tnote{2}          & 41                 & 1,055            & 2               \\
        Sensit\tnote{1}          & 100                & 78,823           & 3               \\
        Usps\tnote{1}            & 256                & 7,291            & 10              \\
        Books\tnote{3}           & 400                & 2,000            & 2               \\
        DVD\tnote{3}             & 400                & 1,999            & 2               \\
        ELEC\tnote{3}            & 400                & 1,998            & 2               \\
        Kitchen\tnote{3}         & 400                & 1,999            & 2               \\
        Isolet\tnote{4}          & 617                & 1,560            & 26              \\
        MNIST\tnote{5}           & 784                & 60,000           & 10              \\
        FashionMNIST\tnote{6}    & 784                & 60,000           & 10              \\
        Colon\tnote{7}           & 2,000               & 62              & 2               \\
        Gisette\tnote{7}         & 4,955               & 5,997            & 2               \\\hline
\end{tabular}
    \begin{tablenotes}
    \item[1] https://www.csie.ntu.edu.tw/\%7Ecjlin/libsvmtools/datasets/ multiclass.html
    \item[2] https://archive.ics.uci.edu/ml/datasets/QSAR+biodegradation
    \item[3] https://jmcauley.ucsd.edu/data/amazon/
    \item[4] https://archive.ics.uci.edu/ml/datasets/isolet
    \item[5] http://yann.lecun.com/exdb/mnist/
    \item[6] https://github.com/zalandoresearch/fashion-mnist
    \item[7] https://www.csie.ntu.edu.tw/\%7Ecjlin/libsvmtools/datasets/ binary.html
    \end{tablenotes}
    \end{threeparttable}
    \end{table}

\subsection{Algorithms}\label{ssec:algorithms}

We compared the following five algorithms:
    \begin{itemize}
      \item PCA-TSK: We first perform PCA, and keep only a few components that capture 95\% of the variance, to reduce the dimensionality, and then train a vanilla TSK fuzzy system introduced in Section~\ref{sec:tsk}. The parameter $\sigma$ is initialized following Gaussian distribution $\sigma\sim\mathcal{N}(1, 0.2)$.
      \item TSK-$h$: This is the vanilla TSK fuzzy system introduced in Section~\ref{sec:tsk}. The parameter $\sigma$ is initialized following Gaussian distribution $\sigma\sim\mathcal{N}(h, 0.2)$.  We set $h$ to $\{1,5,10,50\}$ to validate the influence of saturation on the generalization performance.
      \item TSK-BN-UR: This is the TSK-BN-UR algorithm in \cite{cui2020optimize}. The weight for UR is selected by the validation set. The parameter $\sigma$ is initialized following Gaussian distribution $\sigma\sim\mathcal{N}(1, 0.2)$.
      \item LogTSK: TSK with the log-transformed defuzzification introduced in Section~\ref{ssec:logtsk}. The parameter $\sigma$ is initialized following Gaussian distribution $\sigma\sim\mathcal{N}(1, 0.2)$. Other parameters are initialized by the method described in Section~\ref{ssec:tsk}.
      \item HTSK: This is our proposed HTSK in Section~\ref{ssec:htsk}. The parameter $\sigma$ is initialized following Gaussian distribution $\sigma\sim\mathcal{N}(1, 0.2)$.
    \end{itemize}

All parameters except $\sigma$ were initialized as described in Section~\ref{ssec:tsk}. All models were trained using MBGD-based optimization. We used Adam~\cite{kingma2014adam} optimizer. The learning rate was set to 0.01, which was the best learning rate chosen by cross-validation on most datasets. The batch size was set to 2,048 for MNIST and FashionMNIST, and 512 for all other datasets. If the batch size was larger than the total number of samples $N_t$ in the training set, then we set it to $\min(N_t, 60)$. We randomly selected 10\% samples from the training set as the validation set for early-stopping. The maximum number of epochs was set to 200, and the patience of early-stopping was 20. The best model on the validation set was used for testing. We ran all TSK algorithms with the number of rules $R=30$. All algorithms were repeated ten times and the average performance was reported.

Note that the aim of this paper is not to pursue the state-of-the-art performance on each dataset, so we didn't use cross-validation to select the best hyper-parameters on each dataset, such as the number of rules. We only aim to demonstrate why TSK fuzzy systems perform poorly on high-dimensional datasets, and the improvement of HTSK and LogTSK.

\subsection{Generalization Performances}

\begin{figure*}[htpb]\centering
  \includegraphics[width=0.95\textwidth]{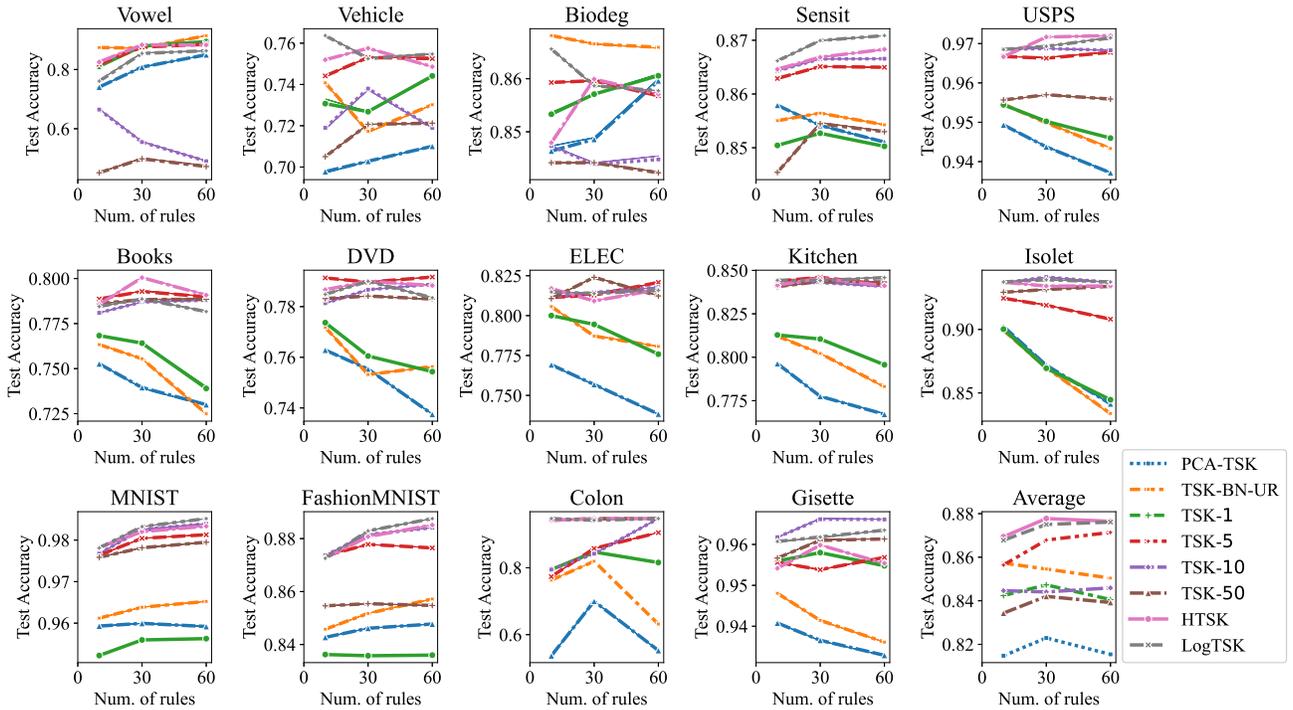}
  \caption{Test accuracies of the eight TSK algorithms on the fourteen datasets with different number of rules.}\label{fig:acc_diff_r}
\end{figure*}

The average test accuracies of the eight TSK algorithms with 30 rules are shown in Table~\ref{tab:acc_r60}. The best accuracy on each dataset is marked in bold. We can observe that:
   \begin{itemize}
     \item On average, HTSK and LogTSK had similar performance, and both outperformed other TSK algorithms on a large range of dimensionality. TSK-$5$ and TSK-$10$ performed well on datasets within a certain range of dimensionality, but they were not always optimal when the dimensionality changed. For instance, on Colon, $h=50$ was better than $h=5$ or $10$, but on Vowel, $h=1$ or $5$ were better than $h=10$ or $50$. However, HTSK and LogTSK always achieved optimal or close-to-optimal performance on those datasets. The results also indicate that the initialization of $h$ should be correlated with $D$, and $h=\sqrt{D}$ is a robust initialization strategy for datasets with a large range of dimensionality.
     \item PCA-TSK performed the worst, which may be because of the loss of information during dimensionality reduction. It also shows the necessity of training TSK models directly on high-dimensional features.
     \item In \cite{cui2020optimize}, TSK-BN-UR outperformed TSK-$1$ on low-dimensional datasets, but this paper shows that it does not cope well with high dimensional datasets.
   \end{itemize}
We also show the test accuracies of the eight TSK algorithms with different number of rules in Fig.~\ref{fig:acc_diff_r}. HTSK and LogTSK outperformed other TSK algorithms, regardless of $R$.

\begin{table*}[htpb]\centering
\caption{Average accuracies of the eight TSK algorithms with 30 rules on the fourteen datasets.} \label{tab:acc_r60}
\begin{tabular}{ccccccccc}\hline
             & PCA-TSK & TSK-BN-UR       & TSK-$1$  & TSK-$5$         & TSK-$10$        & TSK-$50$        & LogTSK          & HTSK             \\
\hline
Vowel        & 80.81   & 87.21           & 87.91    & 87.58           & 55.49           & 49.83           & 85.42           & \textbf{88.32}   \\
Vehicle      & 70.28   & 71.73           & 72.68    & 75.31           & 73.80           & 72.07           & 75.25           & \textbf{75.75}   \\
Biodeg       & 84.86   & \textbf{86.66}  & 85.71    & 85.96           & 84.38           & 84.42           & 85.87           & 85.99            \\
Sensit       & 85.42   & 85.64           & 85.27    & 86.51           & 86.65           & 85.45           & \textbf{87.00}  & 86.68            \\
USPS         & 94.37   & 94.98           & 95.02    & 96.63           & 96.87           & 95.70           & 96.93           & \textbf{97.17}   \\
Books        & 73.95   & 75.55           & 76.42    & 79.28           & 78.70           & 78.83           & 78.87           & \textbf{80.05}   \\
DVD          & 75.53   & 75.32           & 76.05    & 78.97           & 78.67           & 78.42           & \textbf{79.00}  & \textbf{79.00}   \\
Elec         & 75.68   & 78.72           & 79.45    & 81.28           & 81.45           & \textbf{82.40}  & 81.38           & 80.92            \\
Kitchen      & 77.75   & 80.22           & 81.05    & \textbf{84.60}  & 84.30           & 84.37           & 84.42           & 84.55            \\
Isolet       & 87.18   & 86.99           & 86.94    & 91.90           & \textbf{94.12}  & 93.14           & 93.91           & 93.42            \\
MNIST        & 96.00   & 96.38           & 95.60    & 98.04           & 98.26           & 97.82           & \textbf{98.32}  & 98.20            \\
FashionMNIST & 84.61   & 85.17           & 83.58    & 87.78           & 88.15           & 85.54           & \textbf{88.29}  & 88.07            \\
Colon        & 70.00   & 81.95           & 84.74    & 85.79           & 84.21           & \textbf{94.74}  & 94.21           & \textbf{94.74}   \\
Gisette      & 93.66   & 94.14           & 95.80    & 95.38           & \textbf{96.62}  & 96.10           & 96.18           & 95.92            \\
\hline
Average      & 82.15   & 84.33           & 84.73    & 86.79           & 84.41           & 84.20           & 87.50           & \textbf{87.78}   \\
\hline
\end{tabular}
\end{table*}

\begin{table*}[htpb]\centering
\caption{Accuracy ranks of the eight TSK algorithms with 30 rules on the fourteen datasets.} \label{tab:acc_rank}
\begin{tabular}{ccccccccc}\hline
             & PCA-TSK & TSK-BN-UR & TSK-$1$ & TSK-$5$ & TSK-$10$ & TSK-$50$ & LogTSK & HTSK \\ \hline
Vowel        & 6       & 4         & 2       & 3       & 7        & 8        & 5      & 1    \\
Vehicle      & 8       & 7         & 5       & 2       & 4        & 6        & 3      & 1    \\
Biodeg       & 6       & 1         & 5       & 3       & 8        & 7        & 4      & 2    \\
Sensit       & 7       & 5         & 8       & 4       & 3        & 6        & 1      & 2    \\
USPS         & 8       & 7         & 6       & 4       & 3        & 5        & 2      & 1    \\
Books        & 8       & 7         & 6       & 2       & 5        & 4        & 3      & 1    \\
DVD          & 7       & 8         & 6       & 3       & 4        & 5        & 1      & 1    \\
Elec         & 8       & 7         & 6       & 4       & 2        & 1        & 3      & 5    \\
Kitchen      & 8       & 7         & 6       & 1       & 5        & 4        & 3      & 2    \\
Isolet       & 6       & 7         & 8       & 5       & 1        & 4        & 2      & 3    \\
MNIST        & 7       & 6         & 8       & 4       & 2        & 5        & 1      & 3    \\
FashionMNIST & 7       & 6         & 8       & 4       & 2        & 5        & 1      & 3    \\
Colon        & 8       & 7         & 5       & 4       & 6        & 1        & 3      & 1    \\
Gisette      & 8       & 7         & 5       & 6       & 1        & 3        & 2      & 4    \\ \hline
Average      & 7.3     & 6.1       & 6.0     & 3.5     & 3.8      & 4.6      & 2.4    & \textbf{2.1}  \\ \hline
\end{tabular}
\end{table*}

\subsection{Number of Fired Rules}

\begin{figure}[htpb]\centering
      \subfigure[]{\includegraphics[clip,width=\columnwidth]{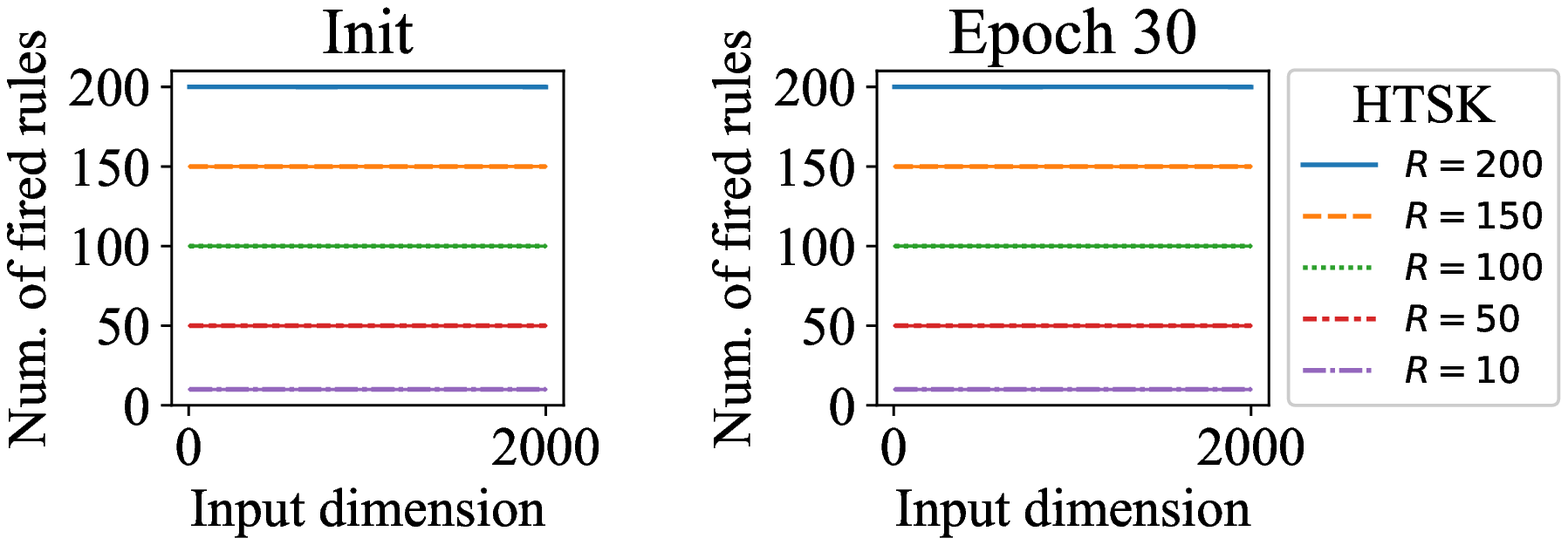}}
      \subfigure[]{\includegraphics[clip,width=\columnwidth]{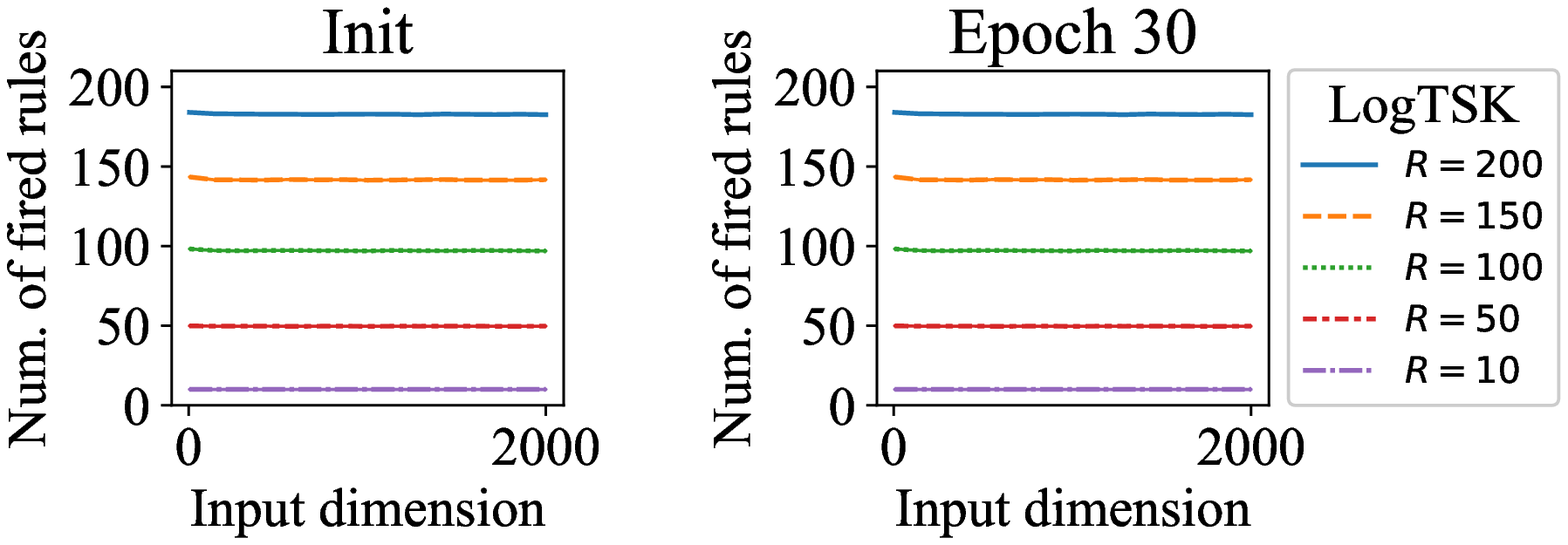}}
      \caption{The number of fired rules by each data point using (a) HTSK, and (b) LogTSK. The first and second columns represent the model before training and after 30 epochs of training, respectively.}\label{fig:dim_frs_htsk_logtsk}
    \end{figure}

We analyzed the number of fired rules by each input on HTSK and LogTSK, and show the results in Fig.~\ref{fig:dim_frs_htsk_logtsk}. The dataset used here is same as the one in Fig.~\ref{fig:dim_frs}. Both figures show that in HTSK and LogTSK, each high-dimensional input fires almost all rules, even with a small initial $\sigma$. However, when the number of rules is large, for instance, $R=200$, LogTSK's number of fired rules is less than 200, but HTSK's number of fired rules is still 200. This may be caused by the $L_1$ normalization of LogTSK, making the normalized firing level sparser than HTSK.

\subsection{Gradient and Loss Landscape}

Figs.~\ref{fig:dim_frs} and \ref{fig:dim_frs_htsk_logtsk} show that $h\geq5$ can counteract most of the influence caused by saturation when $D<1,000$. Therefore, the performances of TSK-$5$, TSK-$10$ and TSK-$50$ are very similar to HTSK and LogTSK on datasets with dimensionalities in that range.

\begin{figure}[htpb]\centering
  \subfigure[]{\includegraphics[clip,width=0.85\columnwidth]{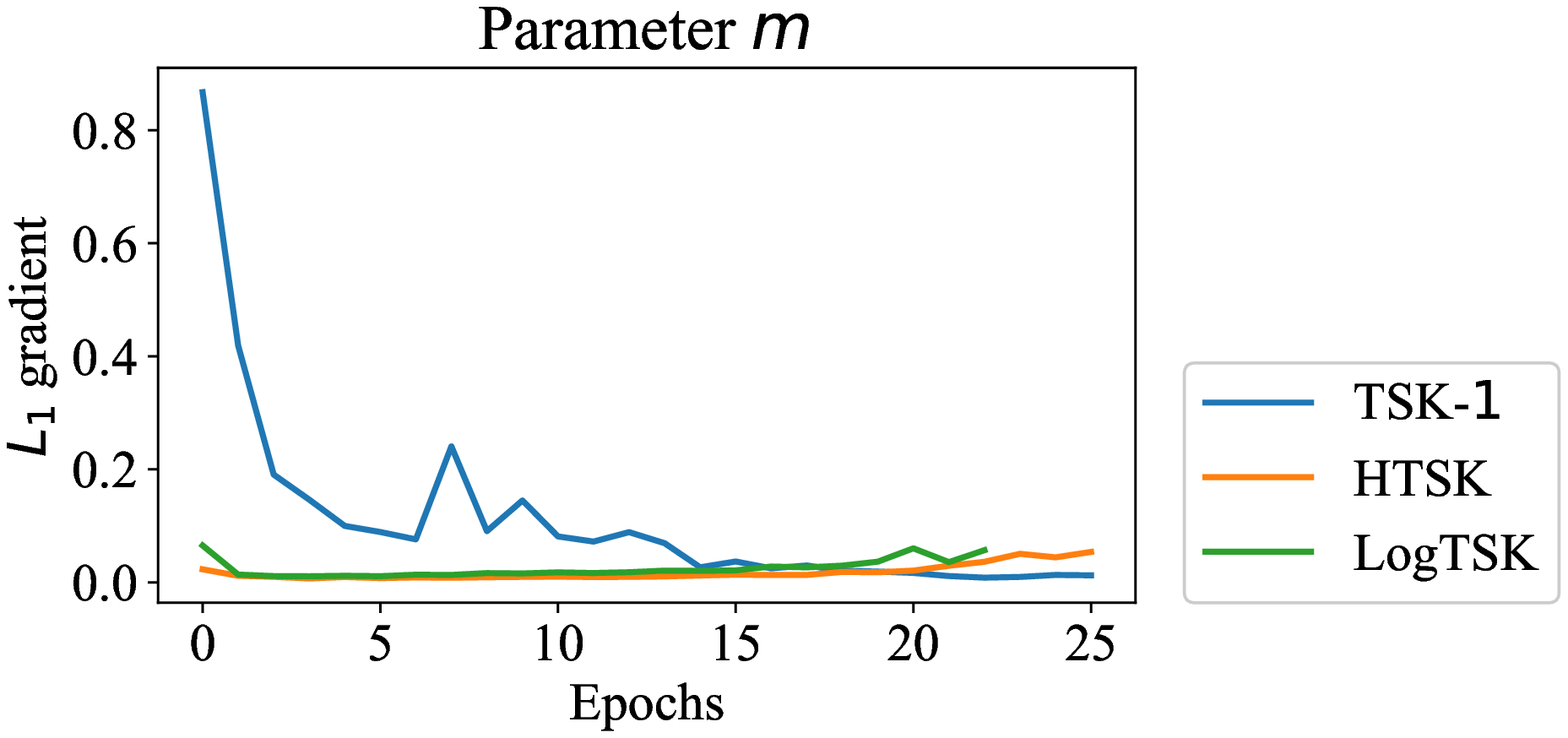}}
  \subfigure[]{\includegraphics[clip,width=0.85\columnwidth]{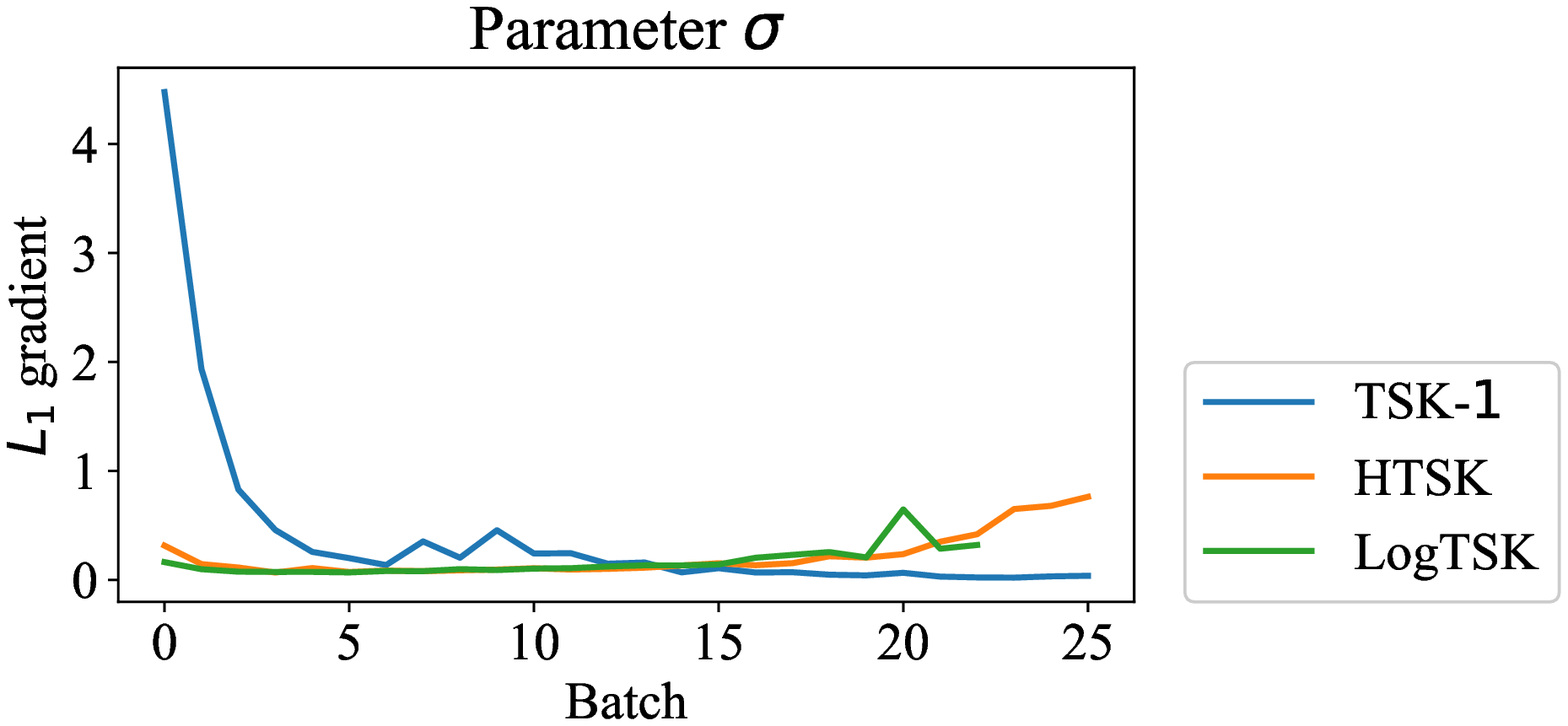}}
  \subfigure[]{\includegraphics[clip,width=0.85\columnwidth]{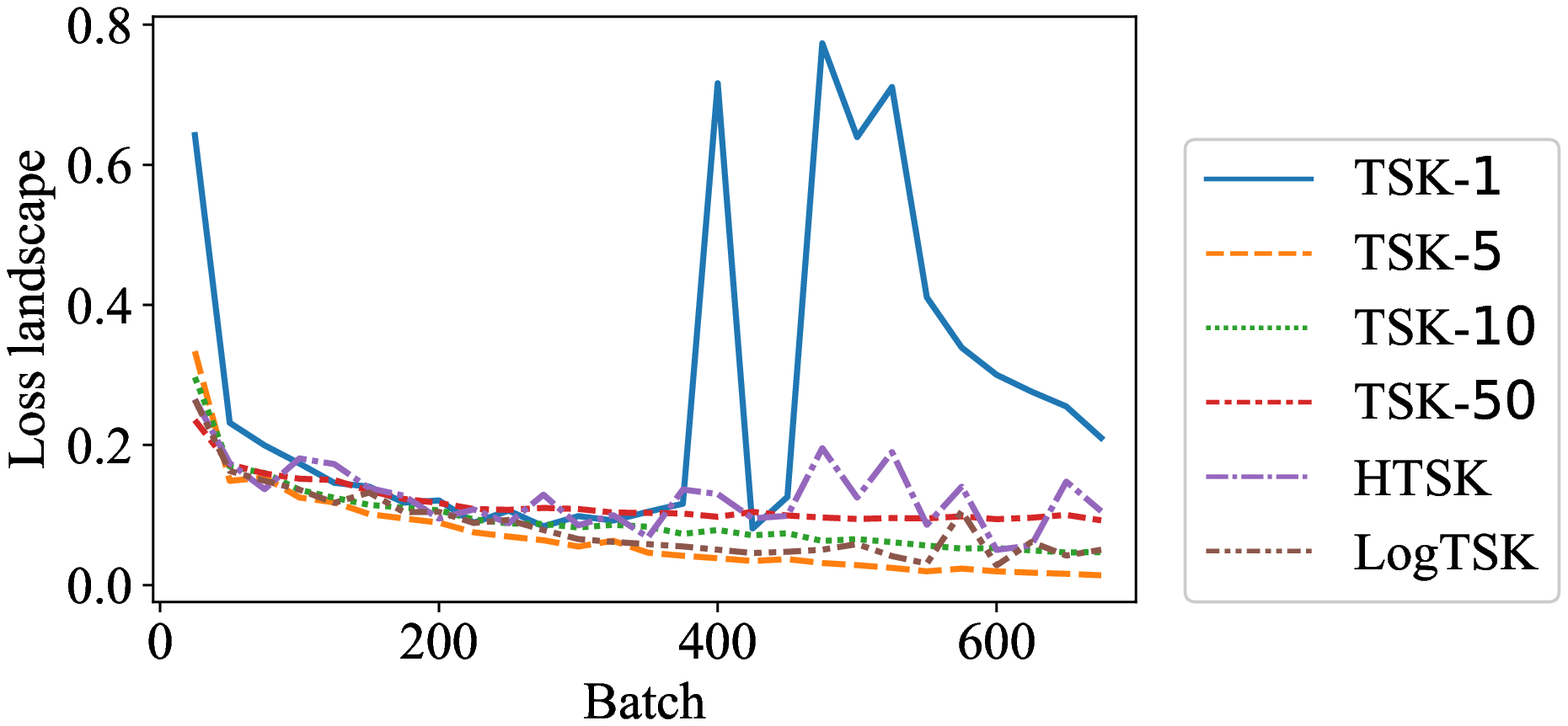}}
  \subfigure[]{\includegraphics[clip,width=0.85\columnwidth]{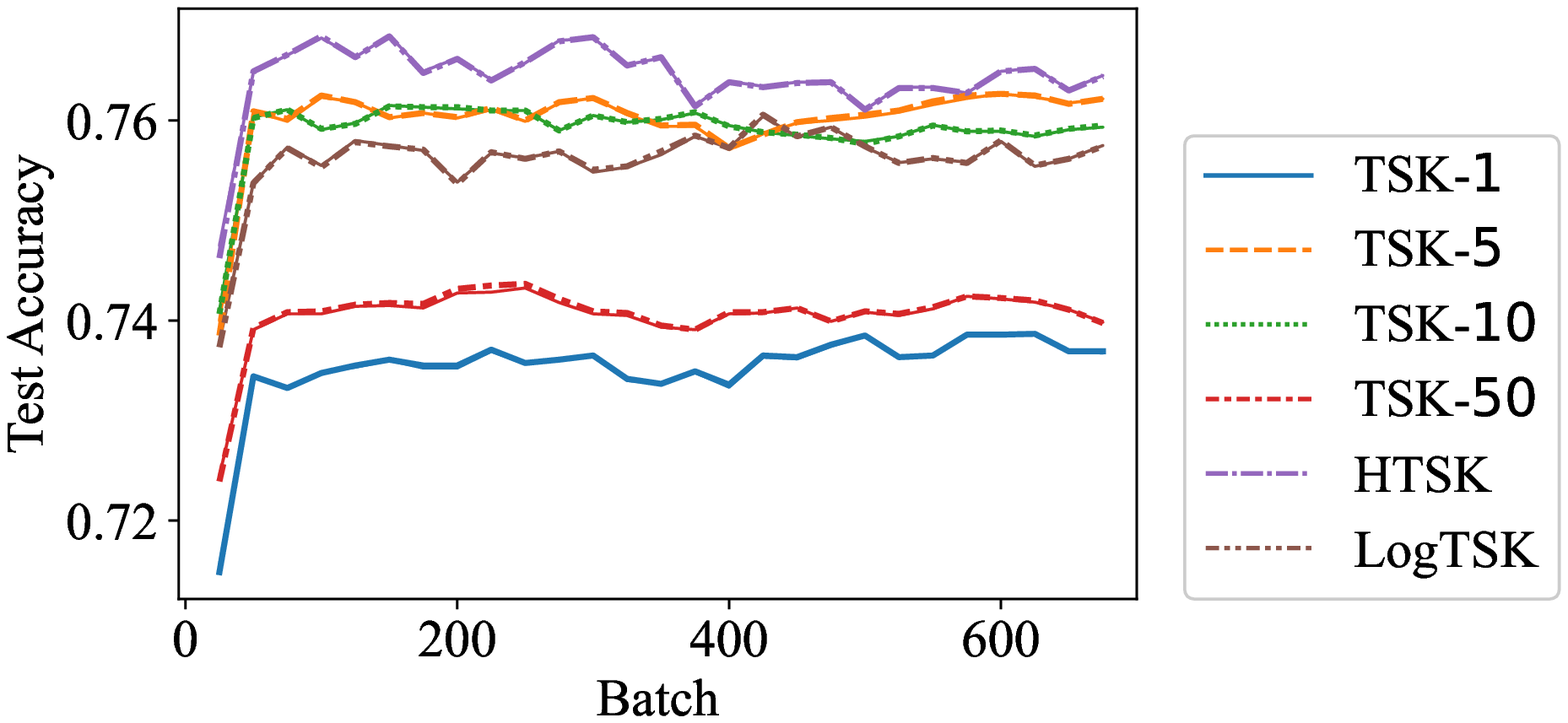}}
  \caption{(a)-(b) The $L_1$ gradients of antecedent parameters $m$ and $\sigma$ versus the training epochs. (c)-(d) The loss landscape and the corresponding test accuracy versus the training batch. All experiments in (a)-(d) were conducted on the Books dataset and repeated ten times.}\label{fig:loss}
\end{figure}

To study if the limited number of fired rules is the only reason causing the decrease of generalization performance, we further analyze the gradient and the loss landscape during training. Because the scale of $\sigma$ affects the gradients' absolute values, we only compare the $L_1$ norm of the gradients for TSK-$1$, HTSK, and LogTSK. The parameter $\sigma$ was initialized following Gaussian distribution $\sigma\sim\mathcal{N}(1, 0.2)$. We recorded the $L_1$ norm of the gradient of the antecedent parameters $m$ and $\sigma$ during training on the Books dataset. Fig.~\ref{fig:loss}(a) and (b) show that the gradient of antecedent parameters from TSK-$1$ is significantly larger than HTSK and LogTSK, especially in the initial training phase.

We also visualize the loss landscape on the gradient direction using the approach in \cite{santurkar2018does}. Specifically, for each update step, we compute the gradient w.r.t. the loss and take one step further using a fixed step: $\eta~\times$ the gradient ($\eta=1$). Then, we record the loss as the parameters move in that direction. When the initial parameters from different runs are the same, the loss' variation represents the smoothness of the loss landscape. Smaller variation means that the loss landscape is flatter, i.e., the gradient would not oscillate when a large learning rate is used, and the gradient is more predictable.

Fig.~\ref{fig:loss}(c) and (d) show the smoothness of the loss landscape and the corresponding test accuracies versus the number of batches. The loss landscapes of the algorithms that can mitigate the saturation are similar, and all are significantly flatter than TSK-$1$. After the model converges on the test accuracy, the variations of the vanilla TSK methods become even larger. This indicates that the gradient of TSK methods is more likely to oscillate during training, which means optimization is more difficult when saturation exists.

The test accuracies of the six TSK algorithms in Fig.~\ref{fig:loss}(d) also demonstrate that, when the loss landscape is more rugged, the generalization performance is worse. Besides, when $h$ is too large, for instance, $h=50$, the generalization performance also decreases. This means finding the proper $h$ is very important, and HTSK and LogTSK should be better choices for training TSK models.

\subsection{Parameter Sensitivity of HTSK and LogTSK}

\begin{figure}[htpb]\centering
    \subfigure[]{\includegraphics[clip,width=0.9\columnwidth]{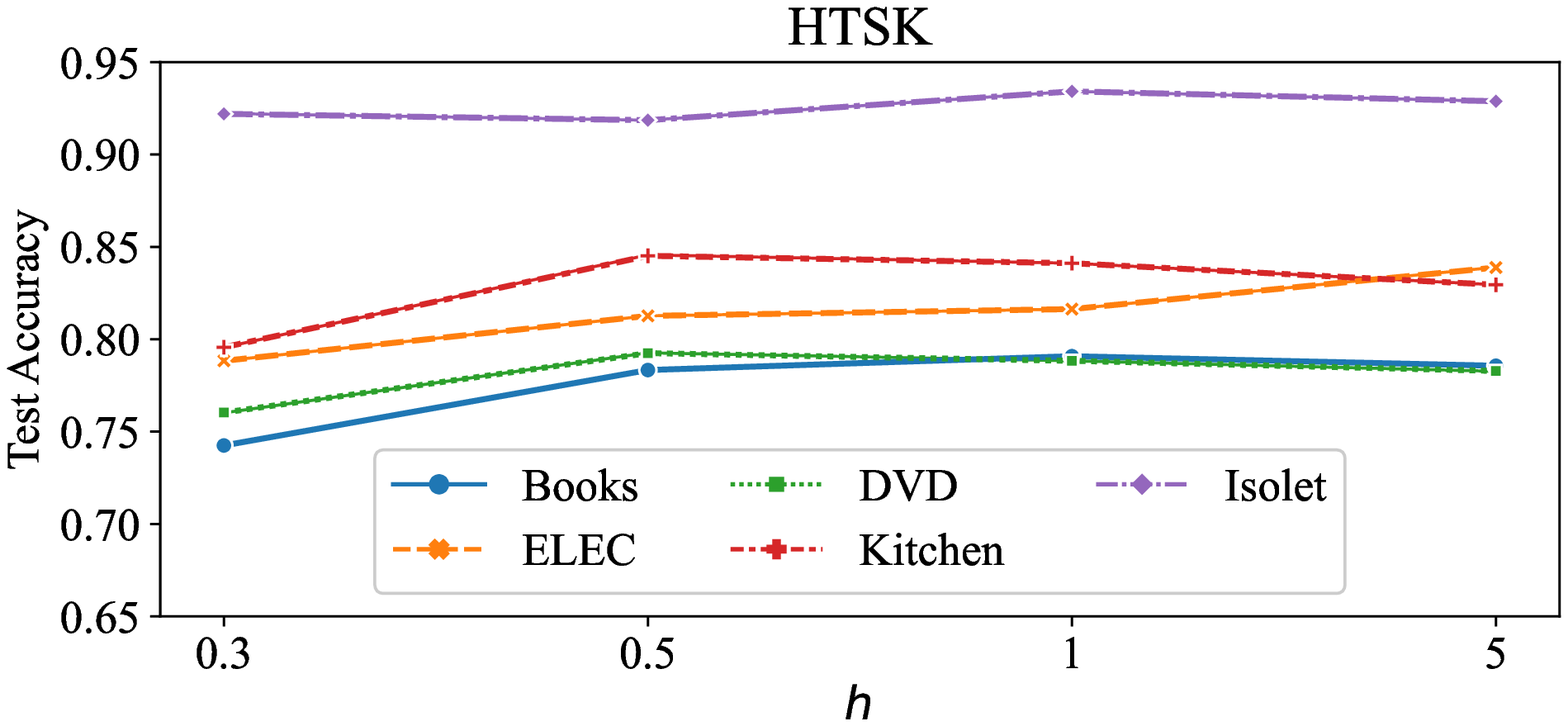}}
    \subfigure[]{\includegraphics[clip,width=0.9\columnwidth]{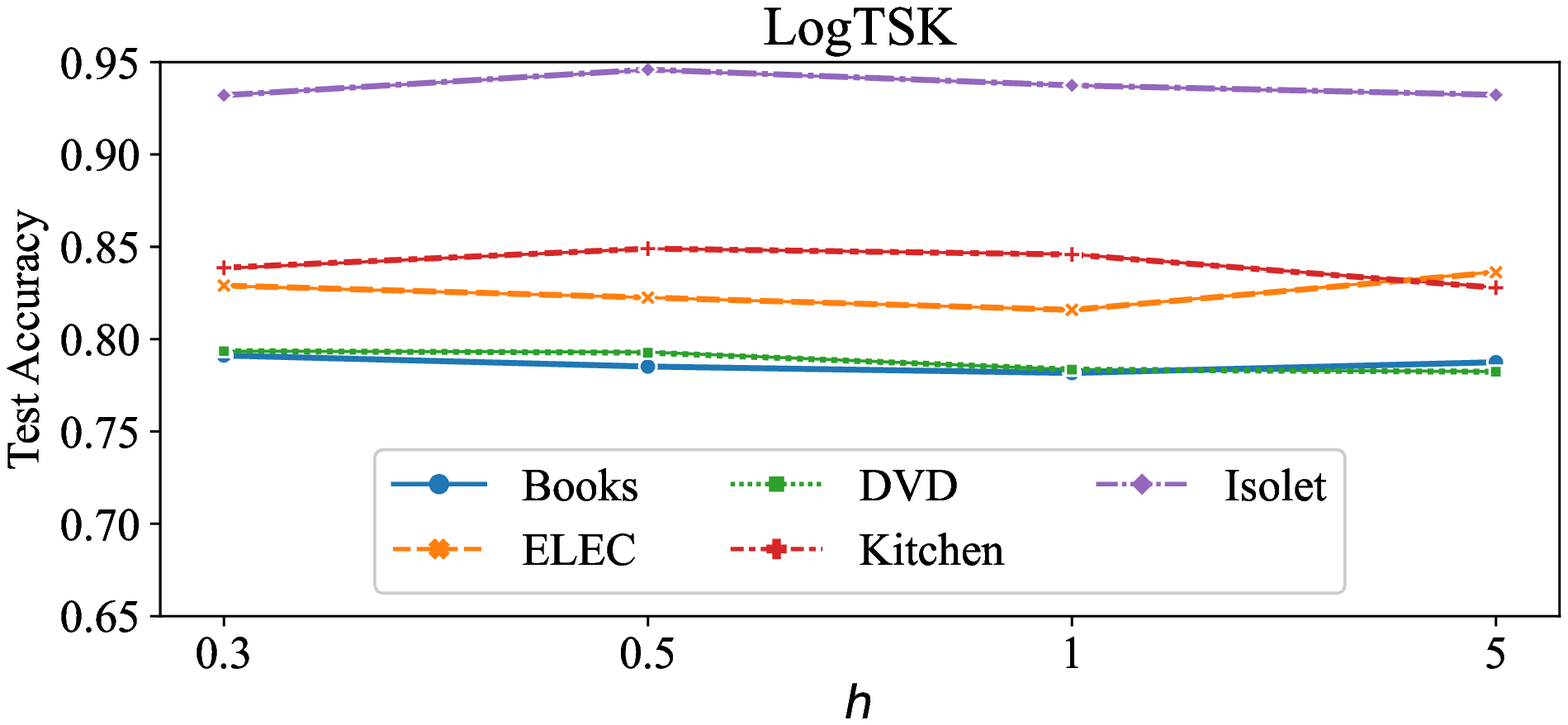}}
    \caption{Test accuracy versus different $h$ of (a) HTSK and (b) LogTSK on five datasets.} \label{fig:acc_diff_h}
  \end{figure}

In the above experiments, we directly set the scale parameter $h=1$ for HTSK and LogTSK. We also compared the generalization performance of HTSK and LogTSK using different $h$. The test accuracies versus different $h$ on five datasets are shown in Fig.~\ref{fig:acc_diff_h}. LogTSK is insensitive to $h$, and HTSK is insensitive to $h$ when $h \geq 0.5$. Smaller $h$ for HTSK leads to larger $|Z_r|$, which may still cause saturation.

In general, $h=1$ is a good choice for both HTSK and LogTSK.

\section{Conclusions}\label{sec:conclusion}

In this paper, we demonstrated that the poor performance of TSK fuzzy systems with Gaussian MFs on high-dimensional datasets is due to the saturation of the \emph{softmax} function. Higher dimensionality causes more severe saturation, making each input fire only very few rules, the gradients of the antecedent parameters become larger and the loss landscape become more rugged. We pointed out that the initialization of $\sigma$ in TSK should be correlated with the input dimensionality to avoid saturation, and proposed HTSK that can handle any-dimensional datasets. We analyzed the performance of two defuzzification algorithms (LogTSK and our proposed HTSK) on datasets with a large range of dimensionality. Experimental results validated that HTSK and LogTSK can reduce the saturation, and both have stable performance.



\end{document}